\documentclass{article}

\usepackage{arxiv}

\usepackage[utf8]{inputenc} 
\usepackage[T1]{fontenc}    
\usepackage{silence}
\usepackage{hyperref}       
\usepackage{url}            
\usepackage{booktabs}       
\usepackage{amsfonts}       
\usepackage{nicefrac}       
\usepackage{microtype}      
\usepackage{lipsum}		
\newif\iffast
\fastfalse
\iffast
\usepackage[draft]{graphicx}
\else
\usepackage{graphicx}
\fi
\usepackage{natbib}

\usepackage{amsmath}
\usepackage{amssymb}
\usepackage{longtable}
\usepackage{multirow}
\usepackage{array}
\usepackage{calc}
\usepackage{listings}
\usepackage[section]{placeins} 

\title{Weak-to-Strong Generalization Enables Fully Automated De
Novo Training of Multi-head Mask-RCNN Model for Segmenting Densely
Overlapping Cell Nuclei in Multiplex Whole-slice Brain Images}

\author{
\normalfont
Lin Bai\textsuperscript{1} \quad
Xiaoyang Li\textsuperscript{1,4} \quad
Liqiang Huang\textsuperscript{1}\\
Quynh Nguyen\textsuperscript{1} \quad
Hien Van Nguyen\textsuperscript{1} \quad
Saurabh Prasad\textsuperscript{1}\\
Dragan Maric\textsuperscript{2} \quad
John Redell\textsuperscript{3} \quad
Pramod Dash\textsuperscript{3}\\
Badrinath Roysam\textsuperscript{1}*
}

\date{
\textsuperscript{1} Department of Electrical and Computer Engineering, University of Houston, TX 77024, USA\\
\textsuperscript{2} National Institute of Neurological Disorders and Stroke, Bethesda, MD 20892, USA\\
\textsuperscript{3} Department of Neurobiology and Anatomy, UT McGovern Medical School, Houston, TX, USA\\
\textsuperscript{4} Canva Pty Ltd., Sydney, NSW, Australia\\
*Corresponding author: \href{mailto:broysam@uh.edu}{broysam@uh.edu}\\[1em]
December 11, 2025
}

\renewcommand{\shorttitle}{Weak-to-Strong Generalization for Nuclear Segmentation}

\hypersetup{
pdftitle={Weak-to-Strong Generalization Enables Fully Automated De Novo Training of Multi-head Mask-RCNN Model for Segmenting Densely Overlapping Cell Nuclei in Multiplex Whole-slice Brain Images},
pdfsubject={q-bio.NC, q-bio.QM},
pdfauthor={Lin Bai, Xiaoyang Li, Liqiang Huang, Quynh Nguyen, Hien Van Nguyen, Saurabh Prasad, Dragan Maric, John Redell, Pramod Dash, Badrinath Roysam},
pdfkeywords={weak to strong learning, automated training, instance segmentation, whole-slide imaging, cell nuclei, multiplex immunofluorescence},
}

\begin{document}
\maketitle

\begin{abstract}
We present a weak to strong generalization methodology for fully automated training of a multi-head extension of the Mask-RCNN method with efficient channel attention for reliable segmentation of overlapping cell nuclei in multiplex cyclic immunofluorescent (IF) whole-slide images (WSI), and present evidence for pseudo-label correction and coverage expansion, the key phenomena underlying weak to strong generalization. This method can learn to segment \emph{de novo} a new class of images from a new instrument and/or a new imaging protocol without the need for human annotations. We also present metrics for automated self-diagnosis of segmentation quality in production environments, where human visual proofreading of massive WSI images is unaffordable. Our method was benchmarked against five current widely used methods and showed a significant improvement. The code, sample WSI images, and high-resolution segmentation results are provided in open form for community adoption and adaptation.
\end{abstract}

\keywords{weak to strong learning \and automated training \and instance segmentation \and whole-slide imaging \and cell nuclei \and multiplex immunofluorescence}

\section{Introduction}

Multiplex immunofluorescent (IF) imaging of whole slides of tissues
using automated cyclic staining is currently performed on a large scale
for deep cellular characterization of histological samples in
pre-clinical science and drug discovery
(\citep{alkofahi2009};
\citep{li2017}; \citep{lin2003}; \citep{lin2007};
\citep{lin2005};
\citep{maric2021};
\citep{rivest2023}). These systems
generate massive multi-gigabyte multi-channel whole-slide images (WSI)
containing hundreds of thousands or more cells that must be analyzed
accurately with maximum-possible and scalable automation. Reliable
detection and accurate segmentation of cell nuclei is a critical yet
challenging first step towards cell-based quantification of molecular
markers. This problem is especially challenging since WSI images cover
extended tissue regions that can exhibit high variability in staining,
tissue architecture, and cell morphology, and varied image artifacts
arising from automated scanning and cyclic immunofluorescence processes.
Interestingly, even the fluorescent staining of cell nuclei is highly
variable across WSI tissue regions and the commonly used DNA stain DAPI
does not mark all nuclei as evident in \textbf{Figure 1A} showing an
image of a rat brain slice ($29,398 \times 43,054$ pixels per channel,
$>$250,000 cells) drawn from a 50-plex dataset
(\citep{maric2021}) (The full
high-resolution image is posted in the Supplement).

Recent progress in segmenting cell nuclei has largely been driven by the
application of deep neural networks and foundational models trained on
massive human-annotated datasets (\citep{kirillov2023}; \citep{radford2021};
\citep{ramesh2022}) in a bid to capture
the variability. Despite the progress, these methods remain human
annotation effort intensive. Foundational models require massive initial
annotation efforts to train, and additional annotation efforts to adapt
the models to a new class of images. Even then, several problems remain.
For example, the correct handling of overlapping nuclei in thicker
samples remains unaddressed (\textbf{Figure 2}). Next, foundational
models are often trained on 3-channel (RGB) images and therefore cannot
exploit the cues that are available in multiplex WSI images, a missed
opportunity (\textbf{Figure 1}). Finally, the problem of assessing the
accuracy of automatically generated segmentations remains human effort
intensive. There is a compelling need for automated assessment methods,
especially in production environments, where visual proofreading of
massive WSI images is unaffordable and not scalable.

We present an integrated \emph{fully automated} methodology based on
\emph{weak-to-strong generalization} (\citep{lang2024}) to overcome the above-mentioned limitations
(\textbf{Figure 3}), and metrics for automated segmentation quality
assessment. Weak-to-strong generalization is a special case of weakly
supervised machine learning in which a strong model (student) is trained
using (comparatively unreliable) annotations generated automatically by
a weaker model (teacher) after they are subjected to well-designed
automated cleanup and data augmentation steps. This strong model is
trained using weak annotations generated using a current foundational
model (\emph{e.g}., SAM (\citep{kirillov2023})) that lacks needed capabilities like the ability to handle
overlapping nuclei. In this work, we refer to these weak annotations as
``pseudo labels'' in line with the weak-to-strong generalization
literature. A key step is the automated processing of the pseudo labels
to generate a large collection of synthetic examples with numerous
instances of complex and variable nuclear overlap patterns. From these
augmented weak label data, we train the strong model by weak-to-strong
learning (\citep{lang2024}). Lang
\emph{et al}. (\citep{lang2024}) showed
that a well-trained student model can learn to correct the weaker
model's mistakes (``pseudo label correction'') and generalize to
instances where the teacher lacks confidence, even when these examples
are excluded from the training data ( ``coverage expansion''). We
present evidence for these phenomena and show their beneficial impact.
Finally, we present metrics for assessing the accuracy of large-scale
nuclear segmentations in a fully automated manner based on accounting
for the fluorescence signal and quantifying the purity of fluorescent
signatures over cells.

\begin{figure}[!htbp]
\centering
\includegraphics[width=\linewidth]{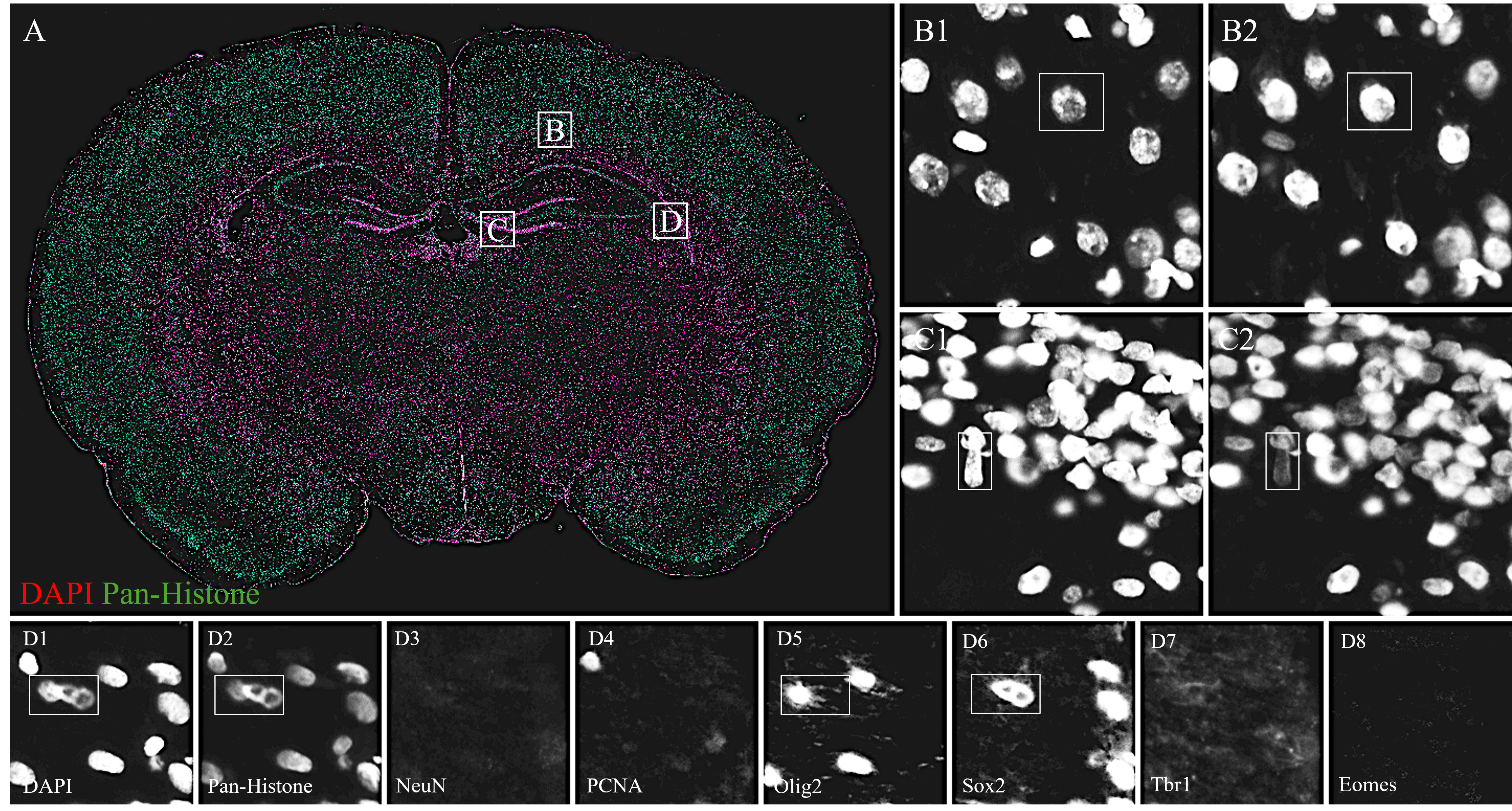}
\caption{\emph{Sample 8-plex image of a whole rat brain slice
($29{,}398 \times 43{,}054$ pixels per channel, $>$250{,}000 cells) showing
major spatial variations in DAPI and histone labeling. (A) RGB rendering
with DAPI in Red, Pan-Histone in Green, and their average in blue. (B1,
B2) Close-up of boxed region B showing strong histone labeling but weak
DAPI labeling. (C1, C2) Close-up of boxed region C showing strong DAPI
labeling but weak histone labeling. (D1--D8) Close-ups of boxed region D
in which we see a cluster of 3 cells that are difficult to separate
based on nuclear stains alone that become possible to distinguish when
cell-type channels Olig2 and Sox2 are utilized.}}
\label{fig:sample_8plex}
\end{figure}

\begin{figure}[!htbp]
\centering
\includegraphics[width=\linewidth]{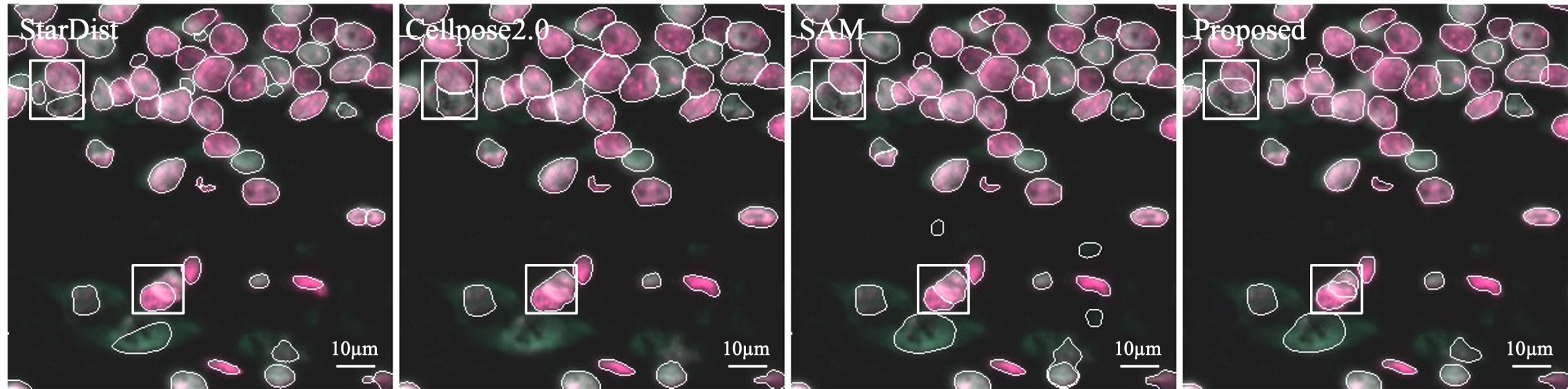}
\caption{\textbf{Figure 2.} \emph{Current methods (StarDist, Cellpose, and SAM)
exhibit poor performance in segmenting overlapping/clustered nuclei. For
example, they either over segment (StarDist) or under segment (Cellpose)
or only able to segment the clearly visible part of nuclei body (SAM).
However, these methods can be used to generate automated weak
annotations to train our strong model that produces accurate results
without requiring any human annotation effort}.}
\label{fig:current_methods}
\end{figure}

\section{Summary of prior work}

Methods for segmenting cell nuclei fall into three main categories.
First, there are parameter-based methods, such as thresholding
(\citep{otsu1975}), watershed
(\citep{malpica1997}), or active
contours (\citep{chan2001}), that
require users to manually adjust parameters for each imaging modality
and scale. This requires care and expertise and often yields suboptimal
performance, largely due to variability across the large WSI images. The
second involves deep CNN models (e.g.,
Mask-RCNN(\citep{he2017}), U-Net
(\citep{ronneberger2015}), YOLACT
(\citep{bolya2019})) that are custom
trained for each class of images to cope with the expected variability.
Variations of these models (\citep{graham2019}; \citep{jia2021};
\citep{johnson2018},
\citep{johnson2020}; \citep{long2020}; \citep{lv2019};
\citep{vuola2019};
\citep{zhou2018}) have been extensively
applied to nuclear segmentation. Deep Learning Models
(\citep{ke2021};
\citep{kong2020};
\citep{molnar2016};
\citep{roy2023};
\citep{zhou2019}) like DoNet
(\citep{jiang2023}) are specifically
designed to identify overlapping objects, and many augmentation methods
(\citep{dvornik2018};
\citep{dwibedi2017};
\citep{fang2019};
\citep{ghiasi2021};
\citep{lin2022};
\citep{yu2023};
\citep{yun2019};
\citep{zhang2017}) are proposed to boost
segmentation performance. While these methods achieve high accuracy,
they require extensive annotation efforts, making them time-consuming
and resource intensive. The third and most scalable approach is to
develop reusable deep learning based models like StarDist
(\citep{schmidt2018}), Cellpose
(\citep{pachitariu2022};
\citep{stringer2025};
\citep{stringer2021}), CellSeg
(\citep{lee2022}) that are pre-trained
on large and diverse datasets. While these models are efficient and
widely used, they often suffer from domain shift -- when the test images
arise from a distribution that is different from the training set, and
the model's performance can degrade. Furthermore, these methods are
designed to work with a small number of channels (e.g., RGB images),
making them sub-optimal in multiplex WSI applications.

Recently, large-scale, pre-trained foundation models (\emph{e.g.} the
Segment Anything Model (SAM) (\citep{kirillov2023})) trained on \textgreater1B masks over 11M images show
competitive segmentation performance and often outperform
fully-supervised models. Foundation models can generalize data
distributions beyond those encountered during training, a capability
facilitated by prompt engineering and referred to as zero-shot and
few-shot generalization. Although powerful, models like SAM cannot
handle overlapping nuclei, and multiplex images since they are trained
on RGB images, and require manual prompts to generate segmentations,
limiting their applicability in fully automated workflows. Many methods
to fine tune SAM for biomedical images have been described, such as
All-in-SAM (\citep{cui2024}), fine-tuning
SAM by providing box level annotations, %
$\mu$SAM
(\citep{archit2025}), fine-tuning SAM for
light and electron microscopy, MedSAM (\citep{ma2024}), fine-tuning SAM on an unprecedented dataset with more
than one million medical image-mask pairs. Unfortunately, fine-tuning
SAM requires a large corpus of (expensive) domain-specific annotation
data.

\begin{figure}[!htbp]
\centering
\includegraphics[width=\linewidth]{figures/image4.png}
\caption{\emph{Illustrating our weak-to-strong learning
framework, where the foundation model SAM serves as a weak teacher to
generate annotations, and our proposed model acts as a strong student to
learn from the generated annotations. (A) First, DAPI and Pan-Histone
images, along with automated point prompts, are input into SAM (weak
teacher) to produce mask annotations. (B) A rule-based filtering process
is then applied to remove union masks and duplicate masks, ensuring
high-quality annotations. (C) The filtered mask annotations, combined
with 8-plex images, are then passed through a data augmentation module,
performing instance-level augmentation. (D) Finally, the augmented data
is used to train our proposed model (strong student) for improved
segmentation post-processing is applied to produce (E)}.}
\label{fig:framework}
\end{figure}

In this paper we show how we can leverage existing models (\emph{e.g.},
SAM) as a teacher model to generate pseudo labels for training a more
capable student model. In weakly supervised learning, for instance,
co-teaching (\citep{han2018}), involves
training two network to teach the other. Co-teaching+
(\citep{yu2019a})builds on this strategy
by selecting low-loss points from samples where the two networks
disagree. Co-learning (\citep{tan2021})
integrates supervised and self-supervised learning with a shared feature
encoder with two exclusive heads, that provide different views of the
data. JoCoR (\citep{zhang2018}) aims to
reduce the diversity of two networks during training. Compared to
classification, fewer studies focus on segmentation tasks. ADELE
(\citep{liu2022}) employs an early
stopping strategy to allow the network to fit clean labels before it
memorizes false annotations. Recently, some authors have developed
theoretic support for weak-to-strong learning
(\citep{lang2024}) for classification
tasks, and identified the conditions under which this strategy is
successful. Our work aims to leverage these emerging insights and adapt
weak to strong learning to instance segmentation.

\section{Weak To Strong Learning Method}

\textbf{Figure 3} depicts the main steps of the proposed weak-to-strong
learning approach. For simplicity, we describe the detailed steps for
fully automated training of a strong model given a single multiplex
image \(I\). This procedure is easily extended to a batch of images. The
trained model can be used to segment similar other images (in inference
mode).

\subsection{Automated Pseudo-label Generation using the Segment Anything Model}

\begin{figure}[!htbp]
\centering
\includegraphics[width=\linewidth]{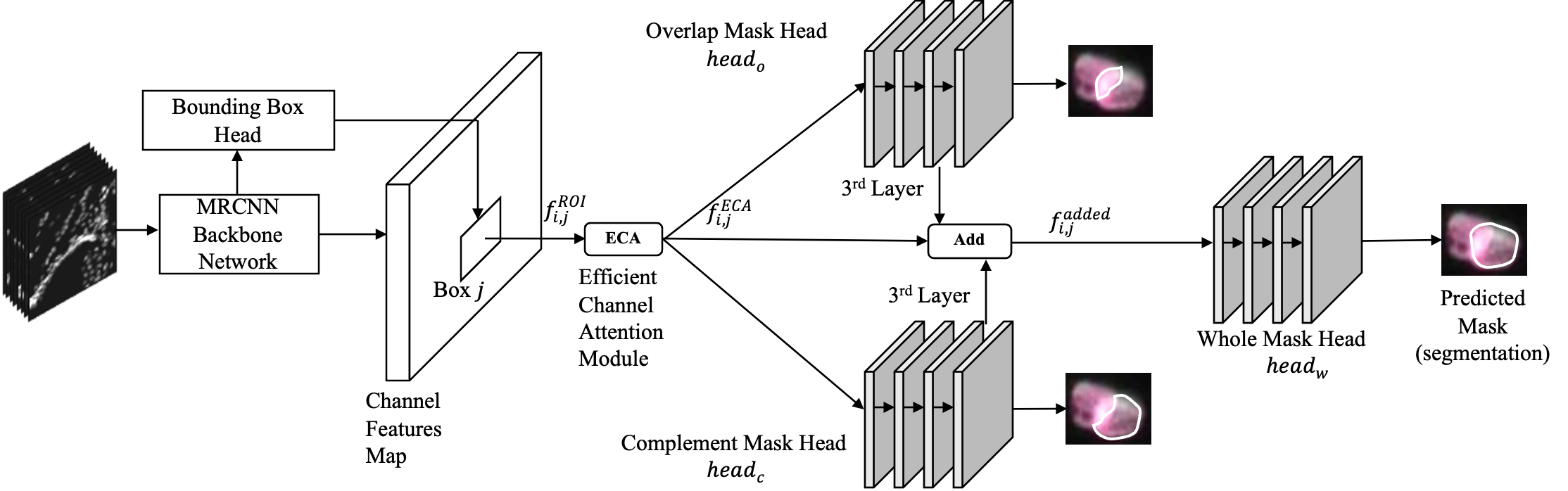}
\caption{\emph{Overview of the proposed M3-RCNN model
architecture. At the detection stage, DAPI and Pan-Histone channels are
used as inputs to generate bounding boxes. At the segmentation stage,
all eight channels are utilized. For each detected box region, the
corresponding channel features are processed through the Efficient
Channel Attention (ECA) module, which dynamically determines the
importance of each channel's contribution to the output. The weighted
features are then passed through a series of convolutional layers to
adjust their feature representations. The adjusted features are
subsequently fed into three segmentation heads: the Overlap Mask Head,
Complement Mask Head, and Whole Mask Head, each predicting different
components of the final mask. The intermediate outputs from the
Intersection Mask Head and Complement Mask Head are added as input to
the Whole Mask Head for enhanced feature learning}.}
\label{fig:m3rcnn_arch}
\end{figure}

Given a large multiplex WSI image \(I\), we dice it into a set
\({\{ x_{i}\}}_{i = 1}^{K}\)of \(K\) 512\(\times\)512-pixel tiles. We
extract a second set of two-channel tiles
\({\{ x_{i}^{DAPI,Histone}\}}_{i = 1}^{K}\) composed of the nuclear
markers DAPI and Pan-histone. From this set, we construct a dataset
\(D = {\{(x_{i}^{DAPI,Histone},\ \ \varepsilon_{i})\}}_{i = 1}^{K}\)
where \(\varepsilon_{i} = {\{ e_{i,j}\}}_{j = 1}^{N_{i}}\) is the set of
masks (pseudo labels) generated by SAM using an array of point prompts,
\(N_{i}\) is the number of masks in the \(i^{th}\) tile, and \(e_{i,j}\)
is the \(j^{th}\) mask in the \(i^{th}\) tile. We also construct a
multiplex version of the dataset
\(D = {\{(x_{i},\varepsilon_{i})\}}_{i = 1}^{K}\). At this stage, SAM
does not identify the categories of the detected objects (\emph{e.g.},
nuclei/other). The masks \(\varepsilon_{i}\) may include noisy labels
due to false detections, over segmentations, under segmentations, or
erroneous overlaps. Accurately identifying these errors requires human
proofreading. However, we can filter the masks considerably using a
simple rule-based strategy focused on two main types of errors: union
masks (two or more smaller masks that are subsumed by a larger mask),
and duplicate masks. The output of the rule-based filtering is
denoted\(\ D^{f} = {\{(x_{i}^{DAPI,Histone},\varepsilon_{i}^{f})\}}_{i = 1}^{K}\)
,
where\(\ \varepsilon_{i}^{f} = {\{ e_{i,j}^{f}\}}_{j = 1}^{N_{i}^{f}}\),
and \(N_{i}^{f}\ \)indicates the revised count of instances in the
\(i^{th}\) image. Details of the rule-based filtering are presented in
pseudocode form as \textbf{Algorithm 1} in the Appendix.

\subsection{Data Augmentation for Simulating Cell Overlaps}

To generate a set of training examples of cell overlaps, we use a
straightforward ``copy and paste'' strategy. Within each image
\(x_{i},\ \)we select two nuclei from\(\ \varepsilon_{i}^{f}\), one to
serve as the `copy' nucleus, which undergoes a combination of spatial
and intensity transformations, and the other as the `paste' nucleus,
onto which the first nucleus is overlayed. Adjustments are made to the
annotations to reflect these changes. Given that
\(\varepsilon_{i}^{f}\)may still include noisy labels, we select the
`copy' and `paste' nuclei based on four criteria: (1) choosing nuclei
that are spatially isolated from others; (2) avoiding nuclei close to
the image boundaries; (3) excluding nuclei with concave mask contours;
and (4) discarding dim nuclei. For the transformation of `copy' nuclei,
we apply random rotation and opacity adjustments to the selected nuclei
to simulate expected overlap patterns. To prevent extreme overlaps, we
implement a threshold on the overlap ratio. The output of these
augmentations is
denoted\(\ D^{aug} = {\{(x_{i}^{aug},\varepsilon_{i}^{aug})\}}_{i = 1}^{K}\),
where\(\ \varepsilon_{i}^{aug} = {\{ e_{i,j}^{aug}\}}_{j = 1}^{N_{i}^{aug}}\),
\(N_{i}^{aug}\ \)represents the revised instance count in the \(i^{th}\)
image. A pseudo-code description is provided as \textbf{Algorithm 2} in
the Appendix.

\subsection{\texorpdfstring{The Proposed Strong Model (M\textsuperscript{3}-RCNN)}{The Proposed Strong Model (M3-RCNN)}}

\textbf{Figure 4} depicts the architecture of the proposed strong
M\textsuperscript{3}-RCNN model. We use the Mask-RCNN
(\citep{he2017}) as the starting point
given its competitive performance in instance segmentation. It operates
in two stages. The first stage employs a backbone network like ResNet-50
(\citep{he2016}) and a feature pyramid
network (FPN) to extract multiscale features of each image tile
\(x^{i}\). A region proposal network (RPN) identifies boxed regions of
interest (ROI), and their features \(f_{i,\ \ j}^{ROI}\). The second
stage consists of a bounding box regression head, classification head
and a mask head that uses the ROI features \(f_{i,\ \ j}^{ROI}\ \)to
generate bounding boxes, object classifications and segmentation masks.
The composite loss function of Mask-RCNN is articulated as:

\begin{equation}
L_{MRCNN} = \lambda_{1}L_{reg} + \lambda_{2}L_{cls} + \lambda_{3}L_{mask}.
\label{eq:lmrcnn}
\end{equation}

where \(L_{reg}\) represents the smooth-L1 loss for bounding box
regression in two stages, \(L_{cls}\) is the cross-entropy loss for
classification in two stages,\(L_{mask}\) is the pixel-wise
cross-entropy loss for segmentation in stage two, and
\(\lambda_{1},\ \lambda_{2}\), \(\lambda_{3}\) are loss weights. Here,
given that we are analyzing eight-channel images, we use an Efficient
Channel Attention (ECA) module (\citep{wang2020}) to weight each channel's feature map to compute the overall
feature map \(f_{i,\ \ j}^{ECA}\) for each box. The ECA module consists
of three components, global average pooling (GAP), one-dimensional
convolution (\(W\)), and a sigmoid function \(\sigma\). The input to the
ECM is \(f_{i,\ \ j}^{ROI}\) for the \(j^{th}\) object from \(i^{th}\)
image tile, and the output \(f_{i,\ \ j}^{ECA}\) is given by:

\begin{equation}
f_{i,j}^{ECA} = \sigma\left( W * GAP\left(f_{i,j}^{ROI}\right) \right).
\label{eq:eca}
\end{equation}

Since nuclei can often overlap, traditional amodal segmentation
approaches concentrate on the visible segments of objects, largely due
to the inherent challenge of representing occluded areas where they are
not directly observable. However, in our context, the objects are not
fully occluded, and some visual cues of the occluded nuclei remain
visible. To leverage these weak but important cues, we diverge from the
conventional single-mask-head framework and introduce two additional
mask heads specifically designed to capture the overlapping and
complement parts of the objects, alongside the standard whole object
representation, as illustrated in \textbf{Figure 4}. We denote these
mask heads \(head_{c}\), \(head_{w},\ \ head_{o,}\) for the complement,
whole mask, and overlap mask heads, respectively. As shown in
\textbf{Figure 4}, the inputs of \(head_{c}\) and \(head_{o}\) are
\(f_{i,\ \ j}^{ECA}\), and the output feature maps of the \(head_{c}\)
and \(head_{o}\) from the 3\textsuperscript{rd} convolution layer are
added with \(f_{i,\ \ j}^{ECA}\) to obtain the feature map
\(f_{i,\ \ j}^{added}\) of the whole mask head. The loss functions for
the mask heads are formulated as follows:

\begin{align}
L_{mask}^{nonconcave} &= \frac{1}{K}\sum_{i = 1}^{K}\frac{1}{N_{i}^{aug,nc}}\sum_{j = 1}^{N_{i}^{aug,nc}}(L_{w} + L_{c} + L_{o}),
\label{eq:lmask_nonconcave}\\
L_{w} &= L_{ce}\left( head_{w}(f_{i,j}^{added}),\ e_{i,j}^{aug} \right),
\label{eq:lw}\\
L_{o} &= L_{ce}\left( head_{o}(f_{i,j}^{ECA}),\ e_{i,j}^{aug,o} \right),
\label{eq:lo}\\
L_{c} &= L_{ce}\left( head_{c}(f_{i,j}^{ECA}),\ e_{i,j}^{aug,c} \right).
\label{eq:lc}
\end{align}

where \(L_{mask}^{non - concave}\) is the loss for non-concave samples
and \(N_{i}^{aug,nc\ }\)represents the total number of instances with
non-concave contours,\(\ e_{i,j}^{aug,\ o}\) and \(e_{i,j}^{aug,\ c}\)
represent the overlapping and complement parts of the augmented
mask\(\ e_{i,j}^{aug}\), respectively, and \(L_{ce}\) is the
cross-entropy loss function. Given two overlapping nuclei A and B, the
segmentation from SAM generally yield a comprehensive mask for nucleus A
and a complementary mask for nucleus B. When analyzing its segmentation
output, denoted as\(\ e_{i,j}^{aug}\) and armed with the prior
understanding that the typical shape of nuclei is non-concave, we
categorize these masks into concave (caused by overlapping) and
non-concave based on the morphology of their contours. Masks exhibiting
non-concave outlines are considered accurate and can be accepted as is.
Masks with concave or incomplete outlines are excluded from the mask
loss computation due to their structural irregularity. However, these
instances are still utilized in the classification and bounding box
regression losses, as their bounding boxes can remain accurate when
overlap is not severe, despite the mask being incomplete. Our
M\textsuperscript{3}-RCNN model is trained in an end-to-end manner using
a composite loss function which is given by:

\begin{equation}
L_{M3RCNN} = \lambda_{1}L_{reg} + \lambda_{2}L_{cls} + \lambda_{3}L_{mask}^{nonconcave}.
\label{eq:lm3rcnn}
\end{equation}

\section{Experimental Results}

The materials and methods for the multiplex WSI imaging are described in
our earlier paper and the data are publicly available
(\citep{maric2021}). We evaluated the
proposed method against five current methods (\textbf{Table 1, Table 2})
in two different ways: (i) automated (unsupervised) assessment using
novel metrics described below; and (ii) manual (supervised) assessment
against ground truth annotations. The natural near symmetry of the rat
brain provides an efficient and practical way to perform these steps.
First, our proposed M\textsuperscript{3}-RCNN model was trained on
\(K = 2,494\) image tiles of size $512 \times 512$ pixels cropped from the left
side of the first whole rat brain slice image (S1). The unsupervised
assessment was carried out over all the brain slices (S1, \ldots{} S4),
by dividing the WSI images into tiles of size $512 \times 512$ pixels with 50
pixels overlap. For the more effort-intensive supervised assessment, we
randomly selected 100 tiles from the right side of the brain in the S1
dataset, annotated them using the semi-automated online computer vision
annotation tool (CVAT), and used this as the ground truth dataset.

Finally, to show the pseudo-label correction phenomenon under our
weak-to-strong generalization framework, we randomly selected 100 tiles
from our training set (left side of the S1 brain), annotated them using
the CVAT annotation tool, and visualized this phenomenon using t-SNE
multi-dimensional projection (\citep{vandermaaten2008}).

For supervised evaluation, we use Aggregated Jaccard Index plus (AJI+)
(\citep{graham2019}) and Panoptic
Quality (PQ) (\citep{kirillov2019}) as
our primary metrics. AJI+ improves upon the original Aggregated Jaccard
Index (AJI) (\citep{kumar2017}) by
penalizing unmatched ground truth instances. PQ jointly evaluate
detection and segmentation quality in a single metric.

\subsection{Automated (Unsupervised) Segmentation Quality Assessment
Metrics}

\begin{figure}[!htbp]
\centering
\includegraphics[width=\linewidth]{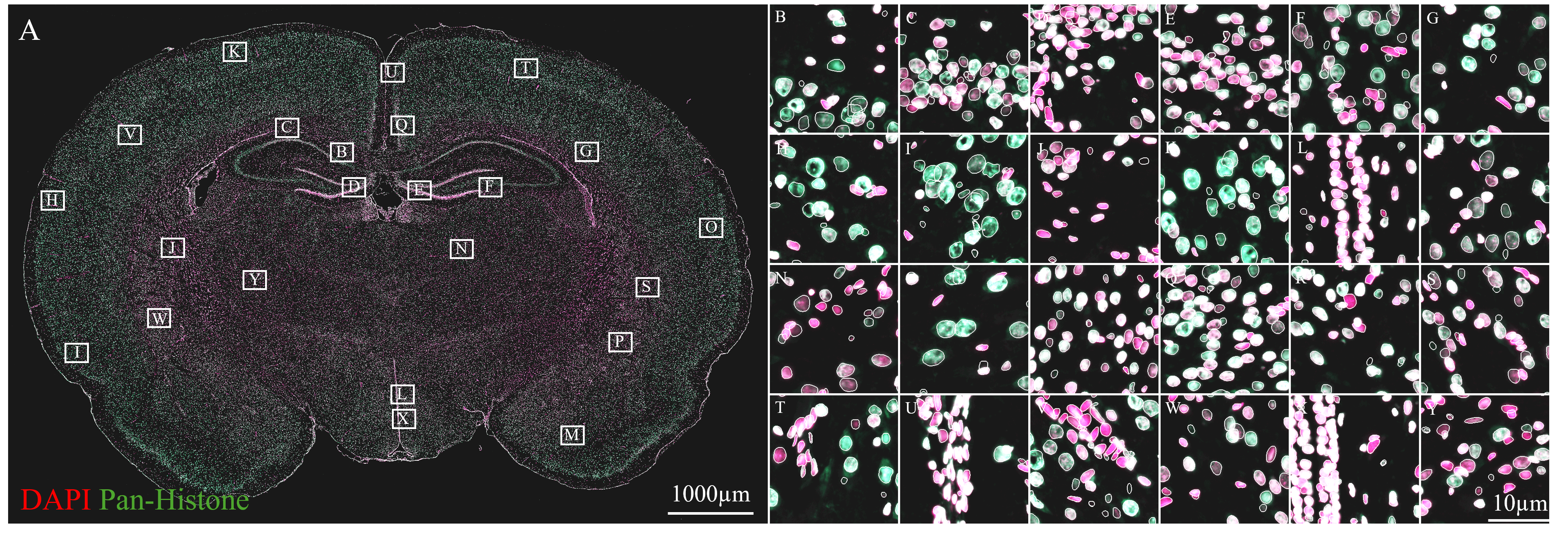}
\caption{\emph{Close-ups of the proposed model's output across
the brain slice, demonstrating its ability to segment overlapping nuclei
despite the variability. The complete high-resolution segmentation files
for four slices S1 -- S4 are provided in the electronic supplement.}
}
\label{fig:qualitative_results}
\end{figure}

As noted earlier, our goal is to assess segmentation quality in a fully
automated manner, without the need for human-generated ground truth
annotations. We recognize that this goal cannot be met for instance
segmentation in general. However, in the context of immunofluorescence
imaging, especially multiplex imaging, it is indeed possible to develop
practically useful quality metrics by examining the extent to which the
fluorescence signal is accounted for by the segmentation masks (``signal
coverage''), and the extent to which each object represents an
individual cell of a certain type as indicated by the expression of a
cell-type marker (we call that marker purity). These metrics are
described below.

\subsubsection{\texorpdfstring{\textbf{(1) Signal Coverage Metric} \(\mathbf{\Gamma}\)}{(1) Signal Coverage Metric (Gamma)}}
\label{signal-coverage-metric-mathbfgamma}

This is a pixel-level measure that quantifies the number of foreground
pixels that are missed by all the segmentation masks generated by the
proposed model, collectively denoted \(M_{model}\). For this, we
identify all the foreground pixels \(M_{DAPI,\ Histone}\) by fitting a
two-component Gaussian Mixture Model (GMM) to separate foreground and
background. With this, the coverage metric \(\Gamma \in (0,1)\) over the
entire image is formulated as:

\begin{equation}
\Gamma = \frac{\sum_{(x,y)} (M_{model} \cap M_{DAPI,\ Histone})}{\sum_{(x,y)} M_{DAPI,\ Histone}}.
\label{eq:coverage_metric}
\end{equation}

where \((x,y)\) denotes all the pixel locations.

\subsubsection{\texorpdfstring{\textbf{(2) Marker Purity Metric} \(\mathbf{\Pi}\)}{(2) Marker Purity Metric (Pi)}}
\label{marker-purity-metric-mathbfpi}

This measure analyzes the expression of cell-type protein markers over
each object, as illustrated in \textbf{Figure 6}. We consider an object
to be `pure' if it contains foreground pixels for one cell type alone,
and `impure' otherwise. To make this notion precise, we model a nuclear
mask \(M_{n}\) as the union of multiple cell-type-specific masks
\(M_{n}^{c}\), with weights \(\alpha_{n}^{c}\  \in (0,1)\), where
\(c = 1,\ ..C\) indexes the cell-type marker channels, and
\(\alpha_{n} = \lbrack\alpha_{n}^{1},\ \ldots\alpha_{n}^{C}\rbrack\) is
the vector of these weights. We perform a sparse decomposition using
Orthogonal Matching Pursuit (OMP) (\citep{pati1993}) to find the most significant cell-type masks and use the
corresponding coefficients \(\alpha_{n}^{c}\) to define the purity
metric for a single mask \(M_{n}\) as follows.

\begin{equation}
{\widehat{\alpha}}_{n} =
\underset{\alpha_{n}}{argmin}
\left\|
ReLuOMP\left(
M_{n} - \sum_{c = 1}^{C}{\alpha_{n}^{c}\ M_{n}^{c}}
\right)
\right\|_{2}^{2}
,\ \ s.t.\ ||\alpha_{n}||_{0} \leq k.
\label{eq:purity_omp}
\end{equation}

where \(k\) is the \(L_{0}\) sparsity constraint (we use \(k = 3)\), and
ReLuOMP is the rectified linear unit function.

With this, the purity metric for the \(n^{th}\) object is given by
\(\Pi_{n} = {\ \max}\left( \alpha_{n}^{1},\ldots\alpha_{n}^{C} \right)\),
and the global purity metric \(\Pi \in (0,1)\) over all \(N\) detected
masks is given by:

\begin{equation}
\Pi = \frac{1}{N}\sum_{n = 1}^{N}\Pi_{n}.
\label{eq:purity_global}
\end{equation}

The
object-wise purity metric \(\Pi_{n}\) can also be used to identify
erroneously segmented objects for corrective editing if desired. For the
cell-type-specific masks \(M_{n}^{c}\), we use a GMM model to identify
the foreground pixels in the cell-type marker channel. In our case, the
biomarkers were NeuN for neurons, Iba-1 for microglia, Olig2 for
oligodendrocytes, S100\\ensuremath{\\beta} for astrocytes and GFP for endothelial cells.
Specifically, we first fit a GMM with three components and then identify
the second largest and largest GMM clusters based on their mean
intensity values. We then calculate the maximum intensity within the
second-largest cluster and the minimum intensity within the largest
cluster. The average of these two values is used as the foreground
threshold. This method is simple yet effective, and filters out the
cytoplasm, processes and auto-fluorescence signals.

\begin{figure}[!htbp]
\centering
\includegraphics[width=\linewidth]{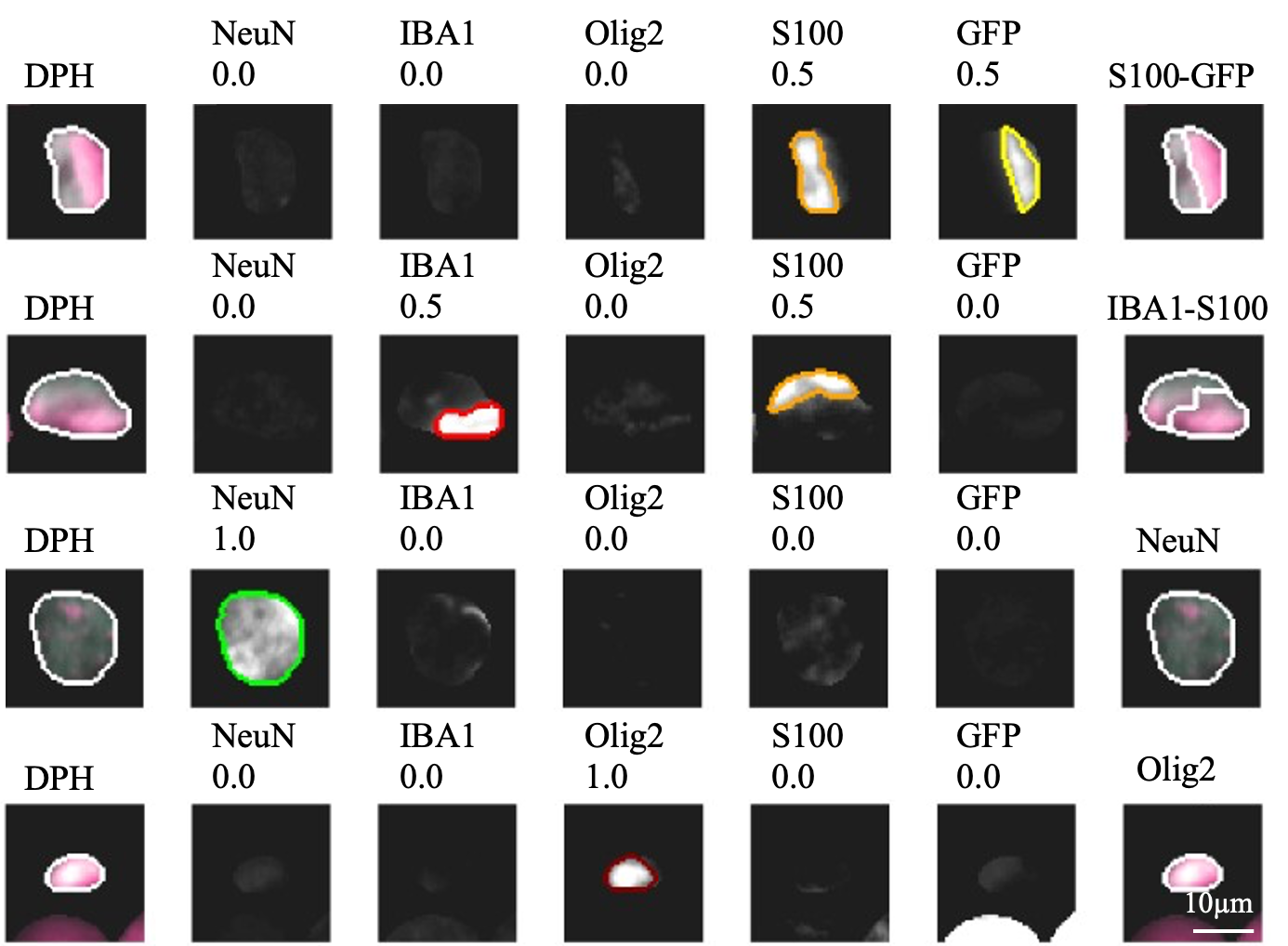}
\caption{\emph{Four examples illustrate the sparse decomposition method for
measuring marker purity for cells. DPH represents DAPI and Pan-Histone.
The first column shows the nuclear segmentations, and the second to
sixth columns show the segmentations of signals in five different
channels containing cell-type markers. The last column shows the final
type. The first row shows an example of a cell with low purity (0.5)
since it contains a mixture of S100 (0.5) and GFP (0.5) signals. The
second row shows an example of a cell with low purity (0.5) since it
contains a mixture of IBA1 (0.5) and S100 (0.5) signals. The third and
fourth rows show high-purity (1.0) examples.}}
\label{fig:marker_purity}
\end{figure}

\subsection{Performance Summary}

\textbf{Table 1} and \textbf{Table 2} provide a performance summary of
the proposed \emph{M\textsuperscript{3}-RCNN} model against five
benchmarks (\emph{Cellpose2.0}, \emph{StarDist}, \emph{SAM-C},
\emph{MRCNN}, \emph{MRCNN-AUG}). For the \emph{Cellpose2.0} and
\emph{StarDist} models, we used their default parameter settings. For
the \emph{SAM-C} method, we provided a $256 \times 256$ array of point prompts to
\emph{SAM} as input and used our rule-based filtering \textbf{Algorithm
1} to eliminate minor errors. For \emph{MRCNN}, we trained the
\emph{MRCNN} model on annotations generated by \emph{SAM}, which were
filtered using \textbf{Algorithm 1}. The \emph{MRCNN-AUG} model was
trained on the annotations generated by \emph{SAM}, which were filtered
using \textbf{Algorithm 1}, and in addition, we applied data
augmentation using \textbf{Algorithm 2}.

\textbf{Table 1: Performance Metrics on the S1 Dataset}

\begin{longtable}[]{@{}
  >{\raggedright\arraybackslash}p{(\columnwidth - 12\tabcolsep) * \real{0.1470}}
  >{\raggedright\arraybackslash}p{(\columnwidth - 12\tabcolsep) * \real{0.1386}}
  >{\raggedright\arraybackslash}p{(\columnwidth - 12\tabcolsep) * \real{0.1386}}
  >{\raggedright\arraybackslash}p{(\columnwidth - 12\tabcolsep) * \real{0.1386}}
  >{\raggedright\arraybackslash}p{(\columnwidth - 12\tabcolsep) * \real{0.1386}}
  >{\raggedright\arraybackslash}p{(\columnwidth - 12\tabcolsep) * \real{0.1386}}
  >{\raggedright\arraybackslash}p{(\columnwidth - 12\tabcolsep) * \real{0.1598}}@{}}
\toprule()
\multirow{2}{*}{\begin{minipage}[b]{\linewidth}\raggedright
Methods
\end{minipage}} &
\multirow{2}{*}{\begin{minipage}[b]{\linewidth}\raggedright
Channels
\end{minipage}} &
\multicolumn{5}{>{\raggedright\arraybackslash}p{(\columnwidth - 12\tabcolsep) * \real{0.7144} + 8\tabcolsep}@{}}{%
\begin{minipage}[b]{\linewidth}\raggedright
S1
\end{minipage}} \\
& & \begin{minipage}[b]{\linewidth}\raggedright
AJI+
\end{minipage} & \begin{minipage}[b]{\linewidth}\raggedright
PQ
\end{minipage} & \begin{minipage}[b]{\linewidth}\raggedright
Coverage
\end{minipage} & \begin{minipage}[b]{\linewidth}\raggedright
Purity
\end{minipage} & \begin{minipage}[b]{\linewidth}\raggedright
Cell Counts
\end{minipage} \\
\midrule()
\endhead
\emph{Cellpose2.0} & 1-plex & 65.6\% & 65.2\% & 82.89\% & 98.99\% &
193,658 \\
\emph{StarDist} & 1-plex & 67.4\% & 68.1\% & 88.47\% & 99.19\% &
247,103 \\
\emph{SAM-C} & 2-plex & 76.0\% & 74.7\% & 97.62\% & 99.12\% & 257,860 \\
\emph{MRCNN} & 2-plex & 76.3\% & 76.0\% & 98.04\% & \textbf{99.30\%} &
\textbf{260,267} \\
\emph{MRCNN-AUG} & 2-plex & 76.2\% & 75.9\% & \textbf{98.26\%} & 99.17\%
& 258,782 \\
\emph{M3-RCNN} & 8-plex & \textbf{77.3\%} & \textbf{76.7\%} & 98.22\% &
99.19\% & 255,726 \\
\bottomrule()
\end{longtable}

\textbf{Table 2: Performance Metrics on the S2, S3, S4 Datasets}

\begin{longtable}[]{@{}
  >{\raggedright\arraybackslash}p{(\columnwidth - 20\tabcolsep) * \real{0.1233}}
  >{\raggedright\arraybackslash}p{(\columnwidth - 20\tabcolsep) * \real{0.0812}}
  >{\raggedright\arraybackslash}p{(\columnwidth - 20\tabcolsep) * \real{0.0856}}
  >{\raggedright\arraybackslash}p{(\columnwidth - 20\tabcolsep) * \real{0.0792}}
  >{\raggedright\arraybackslash}p{(\columnwidth - 20\tabcolsep) * \real{0.0984}}
  >{\raggedright\arraybackslash}p{(\columnwidth - 20\tabcolsep) * \real{0.0856}}
  >{\raggedright\arraybackslash}p{(\columnwidth - 20\tabcolsep) * \real{0.0792}}
  >{\raggedright\arraybackslash}p{(\columnwidth - 20\tabcolsep) * \real{0.0976}}
  >{\raggedright\arraybackslash}p{(\columnwidth - 20\tabcolsep) * \real{0.0856}}
  >{\raggedright\arraybackslash}p{(\columnwidth - 20\tabcolsep) * \real{0.0792}}
  >{\raggedright\arraybackslash}p{(\columnwidth - 20\tabcolsep) * \real{0.1050}}@{}}
\toprule()
\multirow{2}{*}{\begin{minipage}[b]{\linewidth}\raggedright
Methods
\end{minipage}} &
\multirow{2}{*}{\begin{minipage}[b]{\linewidth}\raggedright
Channels
\end{minipage}} &
\multicolumn{3}{>{\raggedright\arraybackslash}p{(\columnwidth - 20\tabcolsep) * \real{0.2632} + 4\tabcolsep}}{%
\begin{minipage}[b]{\linewidth}\raggedright
S2
\end{minipage}} &
\multicolumn{3}{>{\raggedright\arraybackslash}p{(\columnwidth - 20\tabcolsep) * \real{0.2624} + 4\tabcolsep}}{%
\begin{minipage}[b]{\linewidth}\raggedright
S3
\end{minipage}} &
\multicolumn{3}{>{\raggedright\arraybackslash}p{(\columnwidth - 20\tabcolsep) * \real{0.2698} + 4\tabcolsep}@{}}{%
\begin{minipage}[b]{\linewidth}\raggedright
S4
\end{minipage}} \\
& & \begin{minipage}[b]{\linewidth}\raggedright
Coverage
\end{minipage} & \begin{minipage}[b]{\linewidth}\raggedright
Purity
\end{minipage} & \begin{minipage}[b]{\linewidth}\raggedright
Cell Counts
\end{minipage} & \begin{minipage}[b]{\linewidth}\raggedright
Coverage
\end{minipage} & \begin{minipage}[b]{\linewidth}\raggedright
Purity
\end{minipage} & \begin{minipage}[b]{\linewidth}\raggedright
Cell Counts
\end{minipage} & \begin{minipage}[b]{\linewidth}\raggedright
Coverage
\end{minipage} & \begin{minipage}[b]{\linewidth}\raggedright
Purity
\end{minipage} & \begin{minipage}[b]{\linewidth}\raggedright
Cell Counts
\end{minipage} \\
\midrule()
\endhead
\emph{Cellpose2.0} & 1-plex & 74.56\% & 98.57\% & 175,823 & 84.11\% &
99.11\% & 197,136 & 81.17\% & 98.60\% & 193,264 \\
\emph{StarDist} & 1-plex & 93.81\% & 98.94\% & \textbf{250,311} &
88.66\% & 99.40\% & 240,768 & 88.11\% & 99.11\% & 243,169 \\
\emph{SAM-C} & 2-plex & 91.88\% & 98.88\% & 248,263 & 97.65\% & 99.33\%
& \textbf{266,559} & 97.40\% & 99.02\% & \textbf{272,346} \\
\emph{MRCNN} & 2-plex & 93.12\% & \textbf{99.00\%} & 237,956 & 99.05\% &
\textbf{99.40\%} & 252,179 & 98.57\% & \textbf{99.11\%} & 256,516 \\
\emph{MRCNN-AUG} & 2-plex & \textbf{94.26\%} & 98.93\% & 243,830 &
\textbf{99.44\%} & 99.36\% & 257,147 & \textbf{99.00\%} & 99.03\% &
260,994 \\
\emph{M3-RCNN} & 8-plex & 93.93\% & 98.90\% & 241,796 & 98.39\% &
99.39\% & 255,823 & 98.61\% & 99.01\% & 256,789 \\
\bottomrule()
\end{longtable}

\begin{quote}
\emph{1-plex: DAPI, 2-plex: DAPI, Pan-histone, 8-plex: DAPI,
Pan-histone, NeuN, Olig2, PCNA, Sox2, Tbr1, Eomes}

\emph{SAM-C: SAM-Cleaned (Teacher), MRCNN-AUG: Mask-RCNN + Augmentation}
\end{quote}

The proposed (strong) model was trained using the same annotations as
the \emph{MRCNN-AUG} model. To ensure a fair comparison, we trained
\emph{M\textsuperscript{3}-RCNN}, \emph{MRCNN}, and \emph{MRCNN-AUG} for
the 5000 iterations with a batch size 32,
\(\lambda_{1} = \lambda_{2} = \lambda_{3} = 1,\ \beta_{1} = 0.8\),
\(\beta_{2} = 0.7\), \(\beta_{3} = 0.5\), \(t = 3\), utilizing the same
backbone ResNet-50, model hyperparameters, and post-processing.
Post-processing was performed after supervised evaluation in two stages.
First, false positive detections were removed by cross-referencing the
results with those from \emph{MRCNN}. Then, potentially missed
detections were recovered by incorporating segmentation outputs from
\emph{Cellpose2.0} and \emph{StarDist}. For the S1 dataset, our method
achieves the highest AJI+ and PQ scores. \emph{MRCNN} and
\emph{MRCNN-AUG} both outperform \emph{Cellpose2.0}, \emph{StarDist} and
\emph{SAM-C.} In terms of the purity and coverage metrics for the S1 -
S4 datasets, all three methods including
\emph{M\textsuperscript{3}-RCNN,} \emph{MRCNN} and \emph{MRCNN-AUG}
outperform \emph{Cellpose2.0}, \emph{StarDist} and \emph{SAM-C}. This
trend aligns with the results obtained using supervised metrics, further
validating the reliability of our proposed evaluation metric. We can
also see that our baseline \emph{SAM-C} has surpassed \emph{StarDist}
and \emph{Cellpose2.0} due to its training on a large dataset, giving it
a better generalization ability. It is not surprising that
\emph{Cellpose2.0} and \emph{StarDist} do not perform as well since they
have not been trained on this specific dataset.
\emph{M\textsuperscript{3}-RCNN}, \emph{MRCNN} and \emph{MRCNN-AUG}
outperform \emph{SAM-C} which shows the beneficial effect of the
coverage expansion phenomenon in weak to strong generalization.

\textbf{Table 3: Summary of Results from the Ablation Study}

\begin{longtable}[]{@{}
  >{\raggedright\arraybackslash}p{(\columnwidth - 6\tabcolsep) * \real{0.2088}}
  >{\raggedright\arraybackslash}p{(\columnwidth - 6\tabcolsep) * \real{0.3216}}
  >{\raggedright\arraybackslash}p{(\columnwidth - 6\tabcolsep) * \real{0.2128}}
  >{\raggedright\arraybackslash}p{(\columnwidth - 6\tabcolsep) * \real{0.2568}}@{}}
\toprule()
\multicolumn{2}{@{}>{\raggedright\arraybackslash}p{(\columnwidth - 6\tabcolsep) * \real{0.5303} + 2\tabcolsep}}{%
\begin{minipage}[b]{\linewidth}\raggedright
Components
\end{minipage}} &
\multicolumn{2}{>{\raggedright\arraybackslash}p{(\columnwidth - 6\tabcolsep) * \real{0.4697} + 2\tabcolsep}@{}}{%
\begin{minipage}[b]{\linewidth}\raggedright
S1
\end{minipage}} \\
\midrule()
\endhead
ECA & Multi-heads & AJI+ & PQ \\
$\checkmark$ &  & 77.0\% & 76.3\% \\
 & $\checkmark$ & 77.1\% & 76.5\% \\
$\checkmark$ & $\checkmark$ & \textbf{77.3\%} & \textbf{76.7\%} \\
\bottomrule()
\end{longtable}

\textbf{Table 3} provides an ablation study of the proposed
\emph{M\textsuperscript{3}-RCNN} model. Removing ECA module or
multi-heads causes a decrease in performance, showing that each
component contributes independently. The complete model which combines
both modules achieves the best performance. As shown in \textbf{Figure
5}, our model effectively segments entire nuclei for overlapping cases,
even in the densely packed regions. \emph{Our approach can be viewed as
a fully automated refinement method, enhancing existing segmentation
models without the need for expensive ground truth data for training or
fine-tuning.}

\subsection{Evidence for Weak to Strong Generalization}

\begin{figure}[!htbp]
\centering
\includegraphics[width=7.2in,height=3.90069in]{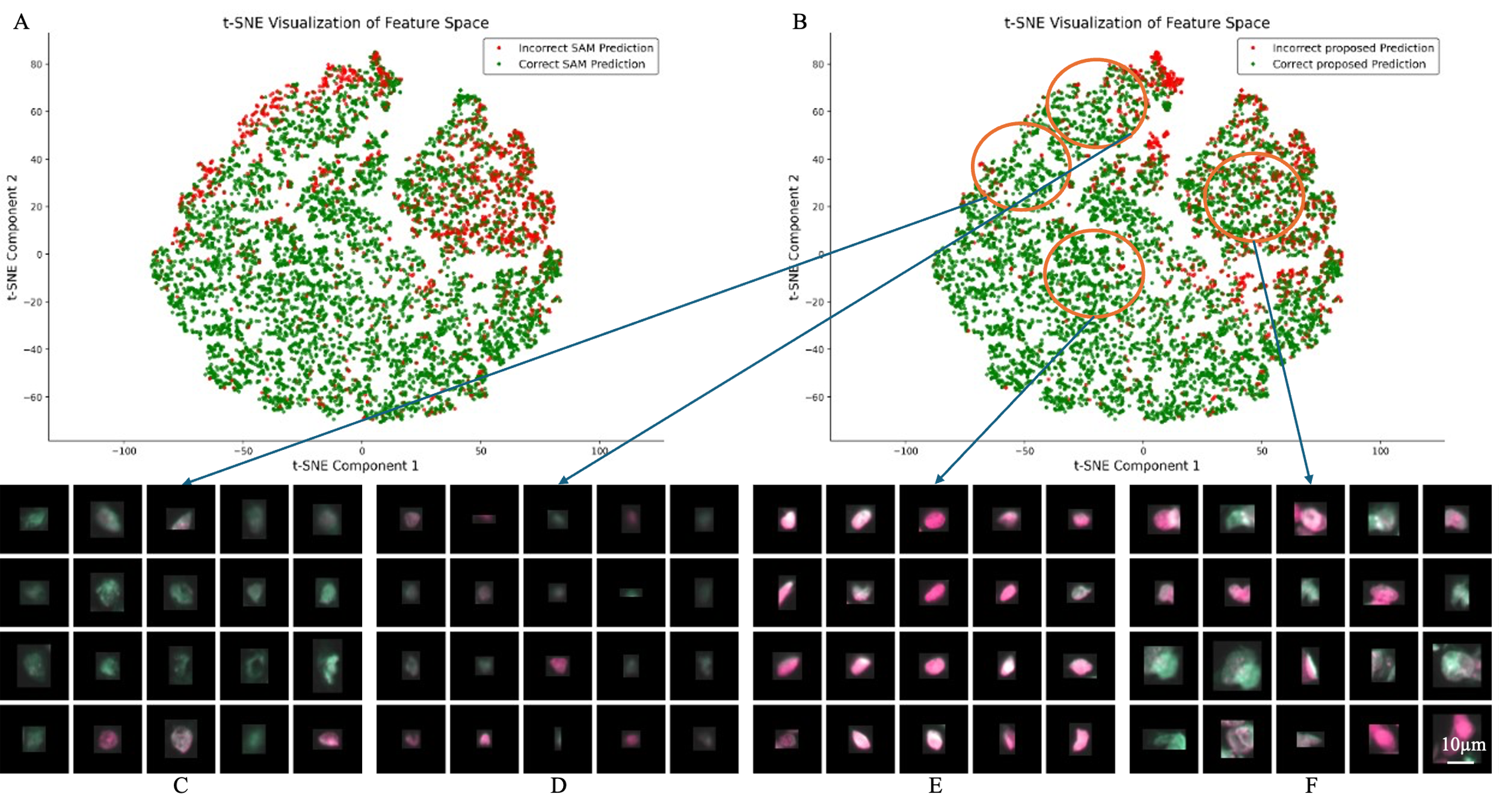}
\caption{\emph{Illustrating the occurrence of pseudo-label
correction in feature space. (A) t-SNE plot with each point corresponding
to a nucleus in the pseudo-label dataset before the weak-to-strong learning, 
where red/green indicate incorrect/correct predictions. 
(B) t-SNE plot after the weak-to-strong learning shows an increase in the 
number of green data points. By visualizing the clusters of data points 
(C--F), we observe improved segmentation of dim nuclei and overlapping nuclei, 
demonstrating the effectiveness of the learning process.}}
\label{fig:pseudolabel_tsne}
\end{figure}

\textbf{Table 4: Measured Expansion and Error Bounds}

\begin{longtable}[]{@{}
  >{\raggedright\arraybackslash}p{(\columnwidth - 12\tabcolsep) * \real{0.1437}}
  >{\raggedright\arraybackslash}p{(\columnwidth - 12\tabcolsep) * \real{0.1700}}
  >{\raggedright\arraybackslash}p{(\columnwidth - 12\tabcolsep) * \real{0.1012}}
  >{\raggedright\arraybackslash}p{(\columnwidth - 12\tabcolsep) * \real{0.1012}}
  >{\raggedright\arraybackslash}p{(\columnwidth - 12\tabcolsep) * \real{0.1826}}
  >{\raggedright\arraybackslash}p{(\columnwidth - 12\tabcolsep) * \real{0.1188}}
  >{\raggedright\arraybackslash}p{(\columnwidth - 12\tabcolsep) * \real{0.1825}}@{}}
\toprule()
\begin{minipage}[b]{\linewidth}\raggedright
Model
\end{minipage} & \begin{minipage}[b]{\linewidth}\raggedright
\[P(\left( \overline{R(f)} \middle| S_{i} \right))\]
\end{minipage} & \begin{minipage}[b]{\linewidth}\raggedright
\[c\]
\end{minipage} & \begin{minipage}[b]{\linewidth}\raggedright
\[\alpha_{i}\]
\end{minipage} & \begin{minipage}[b]{\linewidth}\raggedright
\[err(\left( f,\widetilde{y} \middle| S_{i} \right))\]
\end{minipage} & \begin{minipage}[b]{\linewidth}\raggedright
Bound
\end{minipage} & \begin{minipage}[b]{\linewidth}\raggedright
\[err(\left( f,y \middle| S_{i} \right))\]
\end{minipage} \\
\midrule()
\endhead
\emph{M3-RCNN} & 0.15 & 0.43 & 0.19 & 0.17 & 0.32 & 0.16 \\
\bottomrule()
\end{longtable}

\emph{SAM-C} serves as the weak teacher, and the proposed
\emph{M\textsuperscript{3}-RCNN} serves as the strong student model in
the weak to strong generalization framework. The data augmentation in
\textbf{Algorithm 2} is a crucial link between the weak teacher and the
strong student model. It is clear from \textbf{Table 1} that the
\emph{M\textsuperscript{3}-RCNN} strong student model outperforms the
teacher model on the 100-tiles ground truth dataset that was unseen by
the student model and this is evidence for the coverage expansion
phenomenon (generalizing to instances where the teacher either lacks
confidence or to unlabeled data points that are not part of the training
dataset) we expect in weak to strong generalization.

Next, we compared the pseudo-label data before and after weak-to-strong
learning to examine evidence for the pseudo- label correction phenomenon
(a well-trained student model can learn to correct the weaker model's
mistakes) in feature space (\textbf{Figure 7}). For this, we extracted
all nuclei from 100 labeled images from the training set by locating
their centers and zero padded to $80 \times 80$ -- pixel image patches. Each such
patch (DAPI and Pan-histone) is represented by a seven-dimensional (7-D)
feature vector that quantifies segmentation difficulty in terms of: (i)
foreground contrast; (ii) occlusion score; (iii) boundary variability;
(iv) nucleus size; (v) aspect ratio; (vi) edge intensity; and (vii)
background variability. Foreground contrast measures the difference in
intensity between the nucleus and its background. Occlusion score
evaluates the proportion of a nucleus that is overlapped by other
nuclei. Boundary variability evaluates the number of edge pixels
relative to nucleus size. Nucleus size normalizes the nucleus area by
image size. Aspect Ratio is the width-to-height ratio. Edge intensity is
the mean intensity of nucleus edges. Lastly, Background variability
measures the standard deviation of background pixel intensities.

We used the t-SNE algorithm to project these 7-D features to the plane
for visualization and examined close-ups of image regions for visual
confirmation. In the t-SNE projection, a green dot corresponds to a
correctly predicted nucleus, and a red dot corresponds to an incorrectly
predicted nucleus. To determine whether a prediction is correct (green)
or incorrect (red), we used the following criteria. For non-overlapping
nuclei, a prediction is considered correct if its IoU with the ground
truth exceeds 0.7. For overlapping nuclei, a prediction must cover at
least 10 pixels of the overlapping region and have an IoU \textgreater{}
0.5 prediction to be considered as correct. In comparing \emph{SAM-C}
and our proposed model (\textbf{Figure 7}), we clearly observe a
decrease in the total number of red points indicating the pseudo-label
correction phenomenon, and a visual confirmation of improved
segmentation accuracy. Additionally, by visualizing the clusters of data
points, particularly in regions where pseudo-label correction occurred,
we see that our model achieves better segmentation of large, small and
dim nuclei, as well as nuclei occluded by other nuclei.

Lang \emph{et al}. (\citep{lang2024})
have calculated an upper bound for the error rate
\(err\left( f,y \middle| S_{i} \right)\) of a student model against
ground truth data on a pseudo labeled training subset \(S_{i}\) with
known ground truth. Their estimate was formulated for classification
rather than segmentation problems. With this in mind, we choose to
consider our model \emph{M\textsuperscript{3}-RCNN} as a classifier
\(f\) that classifies image patches, denoted \(x,\) that are centered on
nuclei. Following the notation of Lang \emph{et al}., we denote \(y\) as
the ground truth labeler, and \(\widetilde{y}\) as the pseudo labeler
\emph{SAM-C.} With this
notation,\(\ err\left( f,y \middle| S_{i} \right)\) is the error rate
for \(f\) against the ground truth labels, where \(S_{i}\) represents
the set of extracted image patches around the nuclei (\textbf{Figure
7C-F)}. We use \(S_{i}^{bad}\ \)to denote the incorrectly pseudo labeled
image patches and \(S_{i}^{good}\) to denote the correctly pseudo
labeled patches. With this, the upper bound for
\(err\left( f,y \middle| S_{i} \right)\) is given by:

\begin{equation}
err\left( f,y \middle| S_{i} \right)
\leq
\frac{2\alpha_{i}}{1 - 2\alpha_{i}}
P\left( \overline{R(f)} \middle| S_{i} \right)
err\left( f,\widetilde{y} \middle| S_{i} \right)
\alpha_{i}\left( 1 - \frac{3}{2}c \right).
\label{eq:error_bound}
\end{equation}

where \(\alpha_{i} = P(S_{i}^{bad}|S_{i})\) is the error rate of the
pseudo labeler against ground truth labels, and
\(P\left( \overline{R(f)} \middle| S_{i} \right)\) is a measure of the
classifier \(f\)'s robustness, where a lower value corresponds to a
higher robustness, and \emph{vice versa}. Here,
\(R(f) = \{ x:r(f,x) = 0\},\) where
\(r(f,x) = P\left( f\left( x^{'} \right) \neq f(x) \middle| x^{'}\mathcal{\in N}(x) \right)\)
is the probability that \(f\) assigns different labels to \(x\) and its
neighbor \(x^{'}\) in feature space. In equation (11), \(c\) is the
expansion rate between \(S_{i}^{bad}\) and \(S_{i}^{good}\), which
measure how well two sets are mixed together, and is calculated using
the same formulation described in Lang \emph{et al}.
(\citep{lang2024}). The calculated
values are shown in \textbf{Table 4}. Our model error rate
\(err\left( f,y \middle| S_{i} \right)\) is smaller than \(\alpha_{i}\),
this indicates the occurrence of the pseudo correction phenomenon. Also,
the upper bound of \(err\left( f,y \middle| S_{i} \right)\) equals to
0.32, which is reasonably close to the actual observed value of 0.17,
indicating that the theoretical estimation aligns well with empirical
results.

In summary, we see clear evidence of the two key phenomena associated
with weak to strong generalization -- pseudo-label correction and
coverage expansion. Importantly, these phenomena result in a superior
nuclear segmentation algorithm.

\section{Conclusions and Discussion}

We have demonstrated the practicality of building strong (more capable)
nuclear segmentation models from weak (less capable) models in a fully
automated manner under the framework of weak to strong generalization,
for which evidence is clearly observed. Importantly, the entire process,
starting from the image to segmentation results accompanied by
performance metrics, is fully automated and does not require any manual
annotation effort. \emph{This strategy is made possible by the
availability of versatile and powerful foundational models like SAM, and
the exploitation of image cues that are available in multiplex IF
whole-slide images}. Data cleanup and augmentation algorithms are the
unsung but crucial enabler of weak to strong generalization.

Our method can be used to segment novel image datasets \emph{de novo}
(\emph{e.g.}, images from a new instrument using a new protocol) in a
fully automated manner. The first run of the proposed method on a new
dataset produces two outcomes: (i) accurate fully automated segmentation
results for the new dataset accompanied by automated assessment metrics;
and (ii) a trained \emph{M\textsuperscript{3}-RCNN} model that can be
used for segmenting (in inference mode) a larger collection of similar
datasets (\emph{e.g.}, images from the same instrument using the same
protocols). For whole-slide images (WSIs) with dimensions of
$29,398 \times 43,054$ pixels, our pipeline is capable of generating complete
image predictions within \textasciitilde9 hours (1 min to divide WSIs
into tiles, 4 hours 25 mins to generate masks using SAM, 11 mins to run
\textbf{Algorithm 1}, 2 mins to run \textbf{Algorithm 2,} 2 hours for
training, 1 hour 2 mins for inferencing and postprocessing, 57 mins for
merging results) on a computer with six NVIDIA RTX6000A GPUs, fully
automatically. Once trained, subsequent inferencing using the model
takes \textasciitilde0.01 seconds per tile in the WSI image with 6,012
tiles (\textasciitilde1 hour total).

By introducing ECA module and multi mask heads, our
\emph{M\textsuperscript{3}-RCNN} model learned capabilities that are not
present in the Mask-RCNN model (\emph{e.g.}, improved handling of
overlapped nuclei). The proposed automated (unsupervised) evaluation
metrics offer the ability to assess automated segmentation quality
without the need for human inspection and proofreading. This becomes
more valuable in the current content when automated multiplex IF systems
are routinely producing massive images on a sufficiently large scale
that manual proofreading is unaffordable. When manual editing is needed,
the purity metric \(\Pi_{n}\) can be used to identify the subset of
objects that need corrective editing (efficient analytics-driven
proofreading).

The proposed model is amenable to further refinement and can be adapted
to other image analysis tasks. Going forward, it is possible to develop
even stronger models and update the cleanup and augmentation modules in
concert to develop novel capabilities, with full automation. Even with
brain tissue images, there is an opportunity to refine the method in
extremely dense regions (\emph{e.g.}, the hippocampus). While our data
augmentation algorithm cannot fully simulate the full range of nuclear
overlapping scenarios in such regions, we note that human analysis of
these regions is so challenging that reliable human annotations are
difficult to obtain. Ultimately, the proposed method cannot generalize
to cases that are entirely absent from the training set.

\textbf{Appendix A1: Algorithm 1 (Rule-based filtering)}

\begin{lstlisting}[basicstyle=\ttfamily\footnotesize,frame=single,breaklines=true]
Input:
  D = {(x_i^{DAPI,Histone}, eps_i)}_{i=1..K},  eps_i = {e_{i,j}}_{j=1..N_i}
  parameters: beta1, beta2, beta3
Output:
  D_filtered = {(x_i^{DAPI,Histone}, eps_i_filtered)}_{i=1..K}

for i = 1..K do
    eps_i_u_filtered = {}              # store non-union masks
    for j = 1..N_i do
        eps_i_components = {}          # store component masks
        for m = 1..N_i do
            if j != m and SumPixels(e_{i,j} CAP e_{i,m}) / SumPixels(e_{i,m}) > beta1 then
                add e_{i,m} to eps_i_components
            end if
        end for

        if SumPixels(merge(eps_i_components) CAP e_{i,j}) / SumPixels(e_{i,j}) > beta2
           and count(eps_i_components) > 1 then
            pass                        # union mask (discard)
        else
            add e_{i,j} to eps_i_u_filtered
            # optional: for each component e_{i,m} in eps_i_components:
            #   if isnuclei(x_i^{DAPI,Histone}, e_{i,j} - e_{i,m}) then
            #       add (e_{i,j} - e_{i,m}) to eps_i_u_filtered
            #   end if
        end if
    end for

    eps_i_d_filtered = {}              # store non-duplicate masks
    for p = 1..count(eps_i_u_filtered) do
        for q = 1..count(eps_i_u_filtered) do
            if p != q and
               SumPixels(eps_i_u_filtered[p] CAP eps_i_u_filtered[q]) / SumPixels(eps_i_u_filtered[p]) > beta3 and
               SumPixels(eps_i_u_filtered[p]) / SumPixels(eps_i_u_filtered[q]) < 1 then
                pass                   # duplicate (discard)
            else
                add eps_i_u_filtered[p] to eps_i_d_filtered
                # optional:
                #   if isnuclei(x_i^{DAPI,Histone}, eps_i_u_filtered[p] - eps_i_u_filtered[q]) then
                #       add (eps_i_u_filtered[p] - eps_i_u_filtered[q]) to eps_i_d_filtered
                #   end if
            end if
        end for
    end for

    eps_i_filtered = {}
    for r = 1..count(eps_i_d_filtered) do
        if notdim(x_i^{DAPI,Histone}, eps_i_d_filtered[r]) then
            add eps_i_d_filtered[r] to eps_i_filtered
        end if
    end for
end for
\end{lstlisting}

\textbf{Appendix A2: Algorithm 2 (Data augmentation)}

\begin{lstlisting}[basicstyle=\ttfamily\footnotesize,frame=single,breaklines=true]
Input:
  D_f = {(x_i, eps_i^f)}_{i=1..K}, eps_i^f = {e_{i,j}^f}_{j=1..N_i^f}
  parameter: t
Output:
  D_aug = {(x_i^aug, eps_i^aug)}_{i=1..K}

for i = 1..K do
    obj_i_copy  = eligible_copy(x_i, eps_i^f)
    obj_i_paste = eligible_paste(x_i, eps_i^f)

    for j = 1..count(obj_i_copy) do
        obj_i_trans = augmentation_transform(obj_i_copy[j])
        (x_i, eps_i^f) = copy_paste(obj_i_trans, obj_i_paste, t, x_i, eps_i^f)
    end for

    x_i^aug   = x_i
    eps_i^aug = eps_i^f
end for
\end{lstlisting}

\begin{quote}
\textbf{A3 - Supplemental Files}
\end{quote}

The code, four sample WSI images (S1 -- S4), and full-resolution
segmentation results are provided in open form for viewing, community
adoption, and potential adaptation to other biomedical image analysis
tasks. The WSI images (S1 -- S4) can be found at
\href{https://doi.org/10.6084/m9.figshare.13731585.v1}{figshare}. The
code and results are shared in
\href{https://www.dropbox.com/scl/fo/utd5oxrn83f3qybf3l59y/AFF5w8N-QqiaVGSDOEFfFFg?rlkey=pilymavrojooqyuz23tg5xppj\&st=gk6bn24v\&dl=0}{Dropbox}.

\section*{Acknowledgments}

This work was supported by the National Institutes of Health grant R01NS109118.

\bibliographystyle{unsrtnat}
\bibliography{references}
\end{document}